\documentclass[journal,twoside]{IEEEtran}
\usepackage{cite}
\usepackage[pdftex]{graphicx}
\usepackage{amsmath}
\usepackage[caption=false,font=footnotesize]{subfig}
\usepackage[table,xcdraw, dvipsnames]{xcolor}
\usepackage{xcolor}
\usepackage{multirow}
\usepackage{enumerate}

\usepackage{amssymb}
\usepackage{amsthm}
\usepackage{times}
\usepackage{epsfig}
\usepackage[english]{babel}
\usepackage[utf8]{inputenc}
\usepackage{algorithmic}
\usepackage{algorithm}

\newtheorem{theorem}{Theorem}

\newcommand\Tstrut{\rule{0pt}{2.5ex}}         % = `top' strut
\begin{document}

\hbadness=2000000000
\vbadness=2000000000
\hfuzz=100pt

\setlength{\abovedisplayskip}{2pt}
\setlength{\belowdisplayskip}{2pt}
\setlength{\floatsep}{6pt plus 1.0pt minus 1.0pt}
\setlength{\intextsep}{6pt plus 1.0pt minus 1.0pt}
\setlength{\textfloatsep}{10pt plus 0pt minus 3.0pt}
\setlength{\parskip}{0pt}
\setlength{\abovedisplayshortskip}{0pt}
\setlength{\belowdisplayshortskip}{0pt}

% paper title
\title{Aligning Correlation Information for \\Domain Adaptation in Action Recognition}
% author names and IEEE memberships
% note positions of commas and nonbreaking spaces ( ~ ) LaTeX will not break
% a structure at a ~ so this keeps an author's name from being broken across
% two lines.
% use \thanks{} to gain access to the first footnote area
% a separate \thanks must be used for each paragraph as LaTeX2e's \thanks
% was not built to handle multiple paragraphs
\author{Yuecong~Xu\textsuperscript{\textsection},
        Haozhi~Cao\textsuperscript{\textsection},
        Kezhi~Mao,
        Zhenghua~Chen,
        Lihua~Xie,~\IEEEmembership{Fellow,~IEEE}
        and~Jianfei~Yang\textsuperscript{\textdagger}%
\thanks{Y. Xu is with Institute for Infocomm Research (I\textsuperscript{2}R), Agency for Science, Technology and Research (A*STAR), Singapore.
1 Fusionopolis Way, \#21-01 Connexis, Singapore 138632, Singapore,
(e-mail: xuyu0014@e.ntu.edu.sg).}%
\thanks{H. Cao, K. Mao, L. Xie and J. Yang are with the School of Electrical and Electronic Engineering,
Nanyang Technological University.
50 Nanyang Avenue, Singapore 639798, Singapore,
(e-mail: haozhi001@e.ntu.edu.sg; ekzmao@ntu.edu.sg; yang0478@e.ntu.edu.sg).}%
\thanks{Z. Chen is with Institute for Infocomm Research (I\textsuperscript{2}R) and Centre for Frontier AI Research (CFAR), Agency for Science, Technology and Research (A*STAR), Singapore.
1 Fusionopolis Way, \#21-01 Connexis, Singapore 138632, Singapore,
(e-mail: chen0832@e.ntu.edu.sg).}%
% \thanks{Manuscript received April 19, 2005; revised August 26, 2015.}
}
% The paper headers
\markboth{IEEE Transactions on xxxxxxxxx}%
{Xu \MakeLowercase{\textit{et al.}}: Aligning Correlation Information for Domain Adaptation in Action Recognition}
% use for special paper notices
%\IEEEspecialpapernotice{(Invited Paper)}

% make the title area
\maketitle
\begingroup\renewcommand\thefootnote{\textsection}
\footnotetext{Equal contribution.}
\begingroup\renewcommand\thefootnote{\textdagger}
\footnotetext{Corresponding author.}
\endgroup

%%%%%% Abstract STARTS HERE %%%%%%%%%
% As a general rule, do not put math, special symbols or citations
% in the abstract or keywords.
\begin{abstract}
    Domain adaptation (DA) approaches address domain shift and enable networks to be applied to different scenarios. Although various image DA approaches have been proposed in recent years, there is limited research towards video DA. This is partly due to the complexity in adapting the different modalities of features in videos, which includes the correlation features extracted as long-range dependencies of pixels across spatiotemporal dimensions. The correlation features are highly associated with action classes and proven their effectiveness in accurate video feature extraction through the supervised action recognition task. Yet correlation features of the same action would differ across domains due to domain shift. Therefore we propose a novel Adversarial Correlation Adaptation Network (ACAN) to align action videos by aligning pixel correlations. ACAN aims to minimize the distribution of correlation information, termed as \textit{Pixel Correlation Discrepancy} (PCD). Additionally, video DA research is also limited by the lack of cross-domain video datasets with larger domain shifts. We, therefore, introduce a novel HMDB-ARID dataset with a larger domain shift caused by a larger statistical difference between domains. This dataset is built in an effort to leverage current datasets for dark video classification. Empirical results demonstrate the state-of-the-art performance of our proposed ACAN for both existing and the new video DA datasets.
\end{abstract}

% Note that keywords are not normally used for peer review papers.

% For peer review papers, you can put extra information on the cover
% page as needed:
% \ifCLASSOPTIONpeerreview
% \begin{center} \bfseries EDICS Category: 3-BBND \end{center}
% \fi
% For peerreview papers, this IEEEtran command inserts a page break and
% creates the second title. It will be ignored for other modes.
\IEEEpeerreviewmaketitle

%%%%%% BODY TEXT STARTS HERE %%%%%%%%%
\section{Introduction}
\label{section:intro}

\IEEEPARstart{A}{ction} recognition has long been studied thanks to its applications in various fields. Despite achieving promising results, most research assumes that the distribution of the test data is in line with that of the train data. Meanwhile, due to the high cost of annotating videos, it is desirable if networks trained in one domain could be directly applied to another. However, significant decrease in performances are observed when networks are applied to cross-domain scenarios. To alleviate the impact of domain shift, studies have been conducted on unsupervised domain adaptation (UDA), which aims to leverage data from the labeled source domain to boost performance on the unlabeled target domain~\cite{shao2014transfer,zhao2020review}. Previously, UDA has been mostly explored on image-based tasks, such as image recognition~\cite{ganin2015unsupervised,ma2019deep,kang2020effective}, object detection~\cite{chen2018domain,cai2019exploring,song2020deep} and person re-identification~\cite{yang2020part,ge2020mutual}. 

\begin{figure}[t]
\begin{center}
   \includegraphics[width=.95\linewidth]{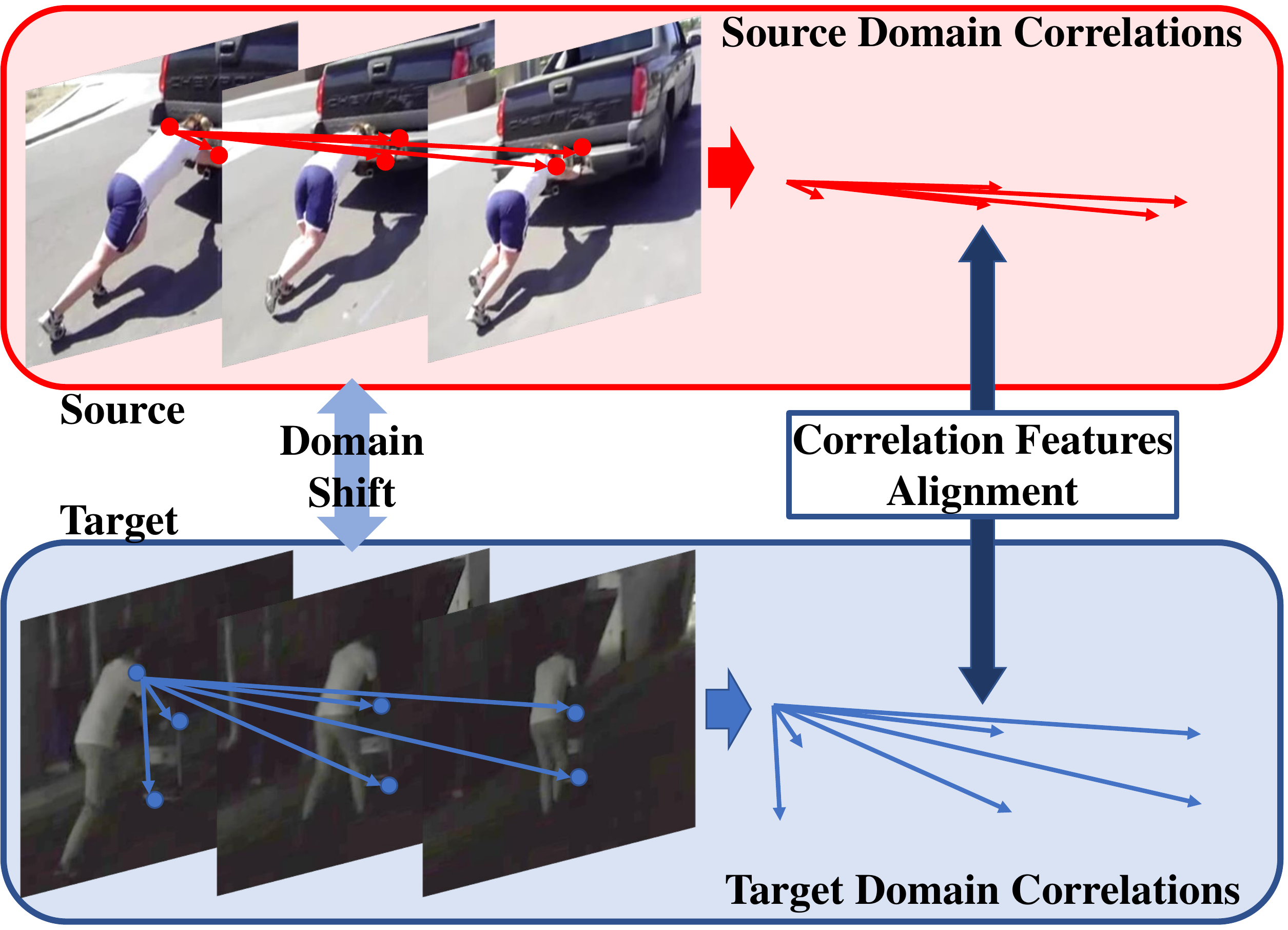}
\end{center}
   \caption{Illustration of our proposed correlation features alignment. The correlation features are extracted as long-range dependencies of pixels across spatiotemporal dimensions. For the same action in the source and target domains, their corresponding correlation features are distinct due to the different postures of the actors. While correlation features are highly associated with the action, alignment of video features should include the alignment of correlation features. Here we show two samples with the action ``Push" from HMDB51 (top) and ARID (bottom).}
\label{figure:1-1-intro}
\end{figure}

Comparatively, there is limited research towards applying DA methods to videos for tasks such as action recognition. This is mainly due to the fact that videos contain data with more modalities, which complicates the adaptation process. Earlier works use the same adaptation strategies as that for image DA while utilizing 3D Convolutional Neural Networks (3D-CNNs) instead of 2D Convolutional Neural Networks (2D-CNNs) for feature extraction. However, these works produce inferior results due to the fact that the simple strategy of substituting feature extractor ignores the different characteristic between spatial and temporal features. Current improvements in DA methods for video tasks focus on improving alignment along the temporal direction. Such improvements are in line with the additional temporal information provided in videos compare to images. They are achieved mainly through applying attention mechanisms to features of video segments sampled across the temporal direction~\cite{chen2019temporal,pan2020adversarial}. Alternatively, auxiliary tasks such as clip order prediction~\cite{xu2019self} are utilized to extract robust temporal representation~\cite{choi2020shuffle}.

Intuitively, the correlation features in videos in the form of long-range spatiotemporal pixel dependencies are highly associated with an action. In supervised action recognition, such correlation features have been recently exploited to aid the extraction of accurate video features. One significant example is the non-local neural network~\cite{wang2018non}, inspired by the non-local mean operation for image denoising \cite{buades2005non,li2016novel}. The spatiotemporal features are constructed by extracting correlation features, obtained by performing self-attention~\cite{vaswani2017attention,chen2020transformer,zhang2020multi}. The correlation features have brought significant increase in network performance compared to utilizing temporal features only~\cite{wang2018non,chen20182,wang2018appearance,yue2018compact,lu2021master}. This is thanks to the fact that temporal features only correlate to local pixel dependencies, while long-range dependencies are captured by correlation features. However, correlation features of the same action could be very different, as depicted in Figure~\ref{figure:1-1-intro}. The same action ``Push" sampled from two different datasets results in distinct correlation information. Given the close relation between correlation features and the action, it is therefore reasonable to not only align spatial and temporal features alone but also to align correlation features. We therefore propose an Adversarial Correlation Adaptation Network (ACAN) that aligns correlation features in an adversarial manner.

For an action within a domain, its correlation features, and the embedded correlation information, would be similar, thanks to the similar appearance and postures of the actors. Yet outliers may be presented in each domain, which may impact the transferability of the network. To cope with such impact, we propose that the joint distribution of correlation information should be aligned. We believe that such a joint distribution of correlation information could be computed as the covariance of the correlation information~\cite{rice2006mathematical}, implemented as its corresponding Gram matrix~\cite{horn2012matrix,ramona2012multiclass}. Therefore, aligning the correlation features of two domains is interpreted as minimizing the difference between the Gram matrices of the correlation information. While direct minimization of the Gram matrix difference could come at a price of decreasing network discriminability and high computation cost, we propose to minimize the \textit{pixel correlation discrepancy} (PCD).

Besides the complexity of the process of video data, the lack of research in DA methods for action recognition and other video-based tasks are also partly due to the lack of sufficient and meaningful cross-domain video datasets. Apart from current video DA datasets, we proposed a new HMDB-ARID dataset from HMDB51~\cite{kuehne2011hmdb} and a recent dark video dataset, ARID~\cite{xu2020arid}. The different illumination conditions of videos in HMDB51 and ARID causes larger domain shift, making the HMDB-ARID dataset more challenging.

Our main contributions are summarized as follows:
\begin{itemize}
    \item We proposed a novel ACAN network for domain adaptation in action recognition by aligning correlation features in the form of long-range spatiotemporal dependencies across domains, which has not been explored by prior works.
    \item We further improve the effectiveness of correlation alignment by aligning the joint distribution of correlation information of different domains through minimizing \textit{pixel correlation discrepancy} (PCD).
    \item We introduce a more challenging video DA dataset: the HMDB-ARID dataset. To our knowledge, this is the first video DA dataset that includes videos shot under different illumination, which possess larger domain shift than current video DA datasets.
    \item We perform extensive experiments, whose results demonstrate the effectiveness of our proposed method, achieving state-of-the-art performance across multiple current and novel video DA datasets.
\end{itemize}

The rest of this paper is organized as follows: related works of unsupervised domain-adaptation in video-based tasks, such as action recognition are discussed in Section~\ref{section:related}. In Section~\ref{section:method}, we introduce our proposed Adversarial Correlation Adaptation Network (ACAN) with the process of minimizing \textit{pixel correlation discrepancy} (PCD) thoroughly. Further, in Section~\ref{section:dataset}, we introduce our proposed HMDB-ARID dataset in detail. After that, we present and analyze the experimental results of our proposed ACAN on previous and our novel video DA datasets, with a thorough ablation study on the design of ACAN in Section~\ref{section:experiments}. Finally, we conclude the paper and propose our future work in Section~\ref{section:conclusion}.

%------------------------------------------------------------------------
\section{Related Works}
\label{section:related}

\subsection{Action Recognition.}
\label{related:ar}
Action recognition has shown great progress with the use of CNNs for extracting accurate video features and representations. There exist mainly two branches of work. One of which utilizes the two-stream structure~\cite{simonyan2014two,feichtenhofer2017spatiotemporal,tran2017two,wang2017two,zhu2018hidden,wang2018temporal,tu2018multi}, extracting video features through CNNs from both optical flow and RGB inputs. The other path utilizes the 3D-CNN structure~\cite{tran2015learning,tran2017convnet,carreira2017quo,liu2018t,hara2018can,yang2019asymmetric,li2020spatio} to extract video features by extracting spatial and temporal features jointly with only RGB inputs. This path has made further progress by introducing separable CNN~\cite{tran2018closer,xie2018rethinking}, improving the efficiency of video feature extraction.

More recently, correlation features in the form of long-range spatiotemporal dependencies have been exploited for further improvements in action recognition. One significant example of which is inspired by the non-local means for image filtering task \cite{buades2005non}, termed the non-local block \cite{wang2018non}, and is introduced with the non-local neural network for capturing correlation between spatiotemporal pixels. Works as in~\cite{ma2018attend,chen20182,zhou2018temporal} also improves video feature extraction using the same idea, but utilizing different methods such as attention~\cite{ma2018attend,chen20182} or relation modules~\cite{zhou2018temporal}. Despite the great progress made in action recognition, most models rely on the target supervised data for fine-tuning on the target dataset, and thus could not be applied to different domains or scenarios without sufficient labels or annotations. To this end, unsupervised domain adaptation helps improve the transferability of models so that they could be applied without access to target labels during training.

\subsection{Unsupervised Domain Adaptation.}
\label{related:uda}
In recent years, there has been a rise of research interest in domain adaptation, which aims to distill shared knowledge across domains and improve the transferability of models. In our work, we focus on unsupervised domain adaptation (UDA), when labeled target data is not available. With the success of Generative Adversarial Network (GAN)~\cite{goodfellow2014generative,liu2019task}, researchers have proposed to construct adversarial loss~\cite{ganin2015unsupervised} for domain adaptation. Various adversarial based domain adaptation methods~\cite{ganin2015unsupervised,tzeng2017adversarial,hoffman2018cycada,zou2019consensus,ma2019deep} have been proposed for a wide range of image-based tasks, such as image recognition~\cite{tzeng2015simultaneous,zhang2017joint,ma2019deep,kang2020effective}, object detection~\cite{chen2018domain,cai2019exploring,zhu2019adapting}, semantic segmentation~\cite{zou2018unsupervised,vu2019dada,chen2019crdoco,guan2021scale} and person re-identification~\cite{panda2019adaptation,yang2020part,ge2020mutual,wang2018learning}.

\begin{figure*}[!ht]
	\begin{center}
		\includegraphics[width=.7\linewidth]{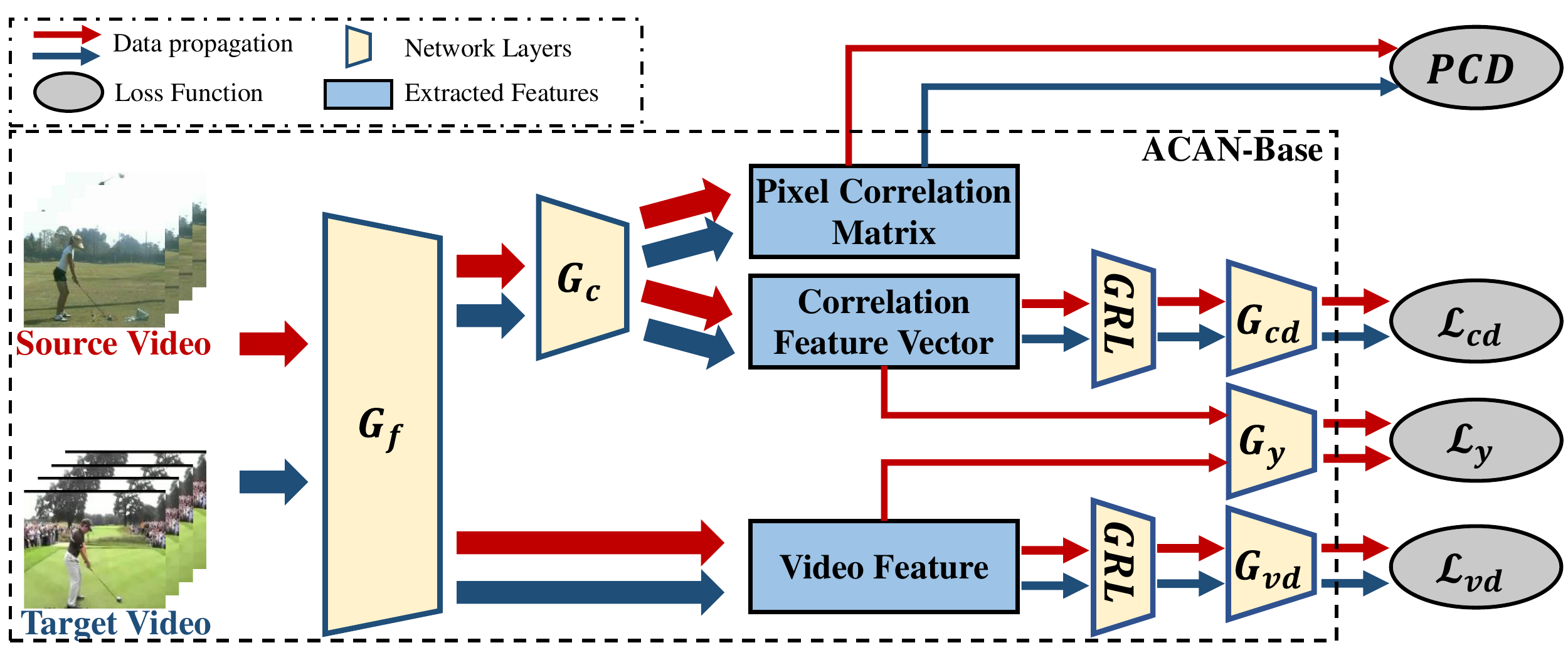}
	\end{center}
	\caption{Overview of the structure of ACAN. We first generate video features with a shared 3D-CNN encoder for both source and target domain videos with the same spatial and temporal dimensions. The source and target correlation feature vectors are obtained through high-level video features, extracted from a deeper layer of the encoder. An adversarial domain loss is applied to both the video features and the correlation feature vectors for aligning the video features and correlation feature vectors. Further, aligning the joint correlation information distribution requires the alignment of the Gram matrices constructed from the pixel correlation matrices (PCM). To achieve this, we further introduce the pixel correlation discrepancy. Figure best viewed in color and zoomed in.}
	\label{figure:3-1-acan}
\end{figure*}

Despite the progress in UDA for image-based tasks, there have been few works on UDA for video-based tasks (VUDA), such as action recognition~\cite{jamal2018deep,chen2019temporal,choi2020shuffle,pan2020adversarial} and action segmentation~\cite{chen2020action}. Compared to direct integration of UDA approaches to videos through a simple change of feature extractor, most of these works adapt temporal features more effectively. However, temporal features only correlate to local pixel dependencies. Meanwhile, none of them have explored the alignment of correlation features that correlate to long-range pixel dependencies, which are highly associated with actions and have proven its effectiveness in supervised tasks, yet may be very different across different domains. We therefore propose to align correlation information for better video feature alignment.

%------------------------------------------------------------------------
\section{Method}
\label{section:method}

In video UDA, we are given a source domain with $N_s$ labeled videos $\mathcal{D}_s=\{(V_s^i,y_s^i)\}^{N_s}_{i=1}$, and a target domain with $N_t$ unlabeled videos $\mathcal{D}_t=\{V_t^j\}^{N_t}_{j=1}$. The source and target domains are characterized by two underlying probability distributions $p_s$ and $p_t$ respectively. The goal of video UDA is to construct a network capable of learning transferable features and minimizing a target classification risk.

Current video DA approaches still rely on aligning only spatial and/or temporal features which correlate local pixel dependencies and fail to align correlation features which correlate long-range pixel dependencies. To cope with this challenge, we propose an Adversarial Correlation Alignment Network (ACAN) to align cross-domain correlation features in an adversarial manner. We further introduce the \textit{pixel correlation discrepancy} (PCD), motivated by the theoretical results in style transfer. We begin this section by presenting the base architecture of ACAN, denoted as ACAN-base, followed by an illustration on the minimization of PCD.

\subsection{Base Architecture}
\label{section:method:arch}
Figure~\ref{figure:3-1-acan} presents the base architecture of our proposed ACAN, illustrated as ACAN-Base. During training, given a source and target video pair $(V_s^i, V_t^j)$, the source and target video features $f_s^i, f_t^j$ are obtained through a shared 3D-CNN encoder $G_f(.;\theta_{f})$. 
% \textcolor{blue}{
To ensure that both the shared encoder is applicable on both the source and target data, the input source and target videos share the same spatial and temporal dimensions. This is achieved by sampling sequentially the same number of frames from both source and target videos, while each frame is resized and cropped directly.
% } 
Meanwhile, the high-level source and target video feature $f_{hs}^i, f_{ht}^j$ are extracted from a deeper layer of $G_f(.;\theta_{f})$ (e.g.\ conv4 layer). The high-level video features are processed by a shared correlation extraction module $G_c$ where the correlation features of the input videos are extracted. The results are the source and target pixel correlation matrices $\mathbf{M}_{s}^i, \mathbf{M}_{t}^j$ as well as the source and target correlation feature vectors $f_{cs}^i, f_{ct}^j$. $G_c(.;\theta_{c})$ is built based on the non-local operation~\cite{wang2018non}, which extracts the correlation features as long-range dependencies between spatiotemporal pixels. To preserve both local and long-range spatiotemporal pixel dependencies, the source correlation feature vector and video feature $f_{cs}^i, f_s^i$ are concatenated to form the overall feature representation of source video $V_s^i$, which would be input to a classifier $G_y$ for action predictions. The action class prediction loss $\mathcal{L}_y$ is computed with respect to the predictions from $G_y$, formulated as:
\begin{equation}
    \mathcal{L}_y = \frac{1}{N_s}\sum\limits_{i=1}^{N_{s}} L_y(G_{y}(f_{cs}^i\oplus f_s^i), y_i),
\end{equation}
where $L_y$ is the cross entropy loss function, and $\oplus$ denotes the concatenation operation.

To accommodate the domain shift between source and target domains, adversarial-based UDA approaches are proved to perform well on image data \cite{ganin2015unsupervised,tzeng2017adversarial,hoffman2018cycada,zou2019consensus} and language data \cite{fu2017domain}. We also leverage such technique for VUDA, which aims to align the global distributions with additional domain discriminators that are trained with the feature generators in a min-max fashion. Domain discriminators are designed to discriminate the video features while the feature generators are trained to deceive the domain discriminators. Here the feature generators are referred to as the combination of $G_f$ and $G_c$. We adopted separate domain discriminators for the source/target video features $f_\ast^\star$ $(\ast\in(s,t),\star\in(i,j))$ and the source/target correlation features $f_{c\ast}^\star$. The two domain discriminators are denoted as the video domain discriminator $G_{vd}(.;\theta_{vd})$ and the correlation domain discriminator $G_{cd}(.;\theta_{cd})$. During the adversarial training process, the parameters $\theta_{vd}$ and $\theta_{cd}$ are learned by minimizing the video domain loss $\mathcal{L}_{vd}$ and the correlation domain loss $\mathcal{L}_{cd}$, respectively, which are formulated as:
\begin{equation}
    \mathcal{L}_{vd} = \frac{1}{N_s}\sum\limits_{i=1}^{N_{s}} L_b(G_{vd}(f_s^i), d_i) + \\ \frac{1}{N_t}\sum\limits_{j=1}^{N_{t}} L_b(G_{vd}(f_t^j), d_j),
\end{equation}
\begin{equation}
    \mathcal{L}_{cd} = \frac{1}{N_s}\sum\limits_{i=1}^{N_{s}} L_b(G_{cd}(f_{cs}^i), d_i) + \\ \frac{1}{N_t}\sum\limits_{j=1}^{N_{t}} L_b(G_{vd}(f_{ct}^j), d_j),
\end{equation}
where $L_b$ is the binary cross-entropy loss of the domain discriminators, while $d_i$ and $d_j$ are the domain label for the source and target domains respectively. Meanwhile, the parameters of the feature extractors $\theta_{f}$ and $\theta_{c}$ are learned to maximize the domain losses simultaneously. To achieve uniform minimization of the action class prediction loss and the maximization of the domain losses, a Gradient Reverse Layer (GRL) \cite{ganin2015unsupervised} is inserted before each domain discriminator as in Figure~\ref{figure:3-1-acan}.

The overall loss function to be optimized can therefore be formulated as:
\begin{equation}
    \mathcal{L} = \mathcal{L}_y - (\lambda_v\mathcal{L}_{vd} + \lambda_r\mathcal{L}_{cd}),
\end{equation}
where $\lambda_v$ and $\lambda_r$ are the trade-off weights for the video domain loss and correlation domain loss respectively.

\begin{figure}[t]
\begin{center}
   \includegraphics[width=.95\linewidth]{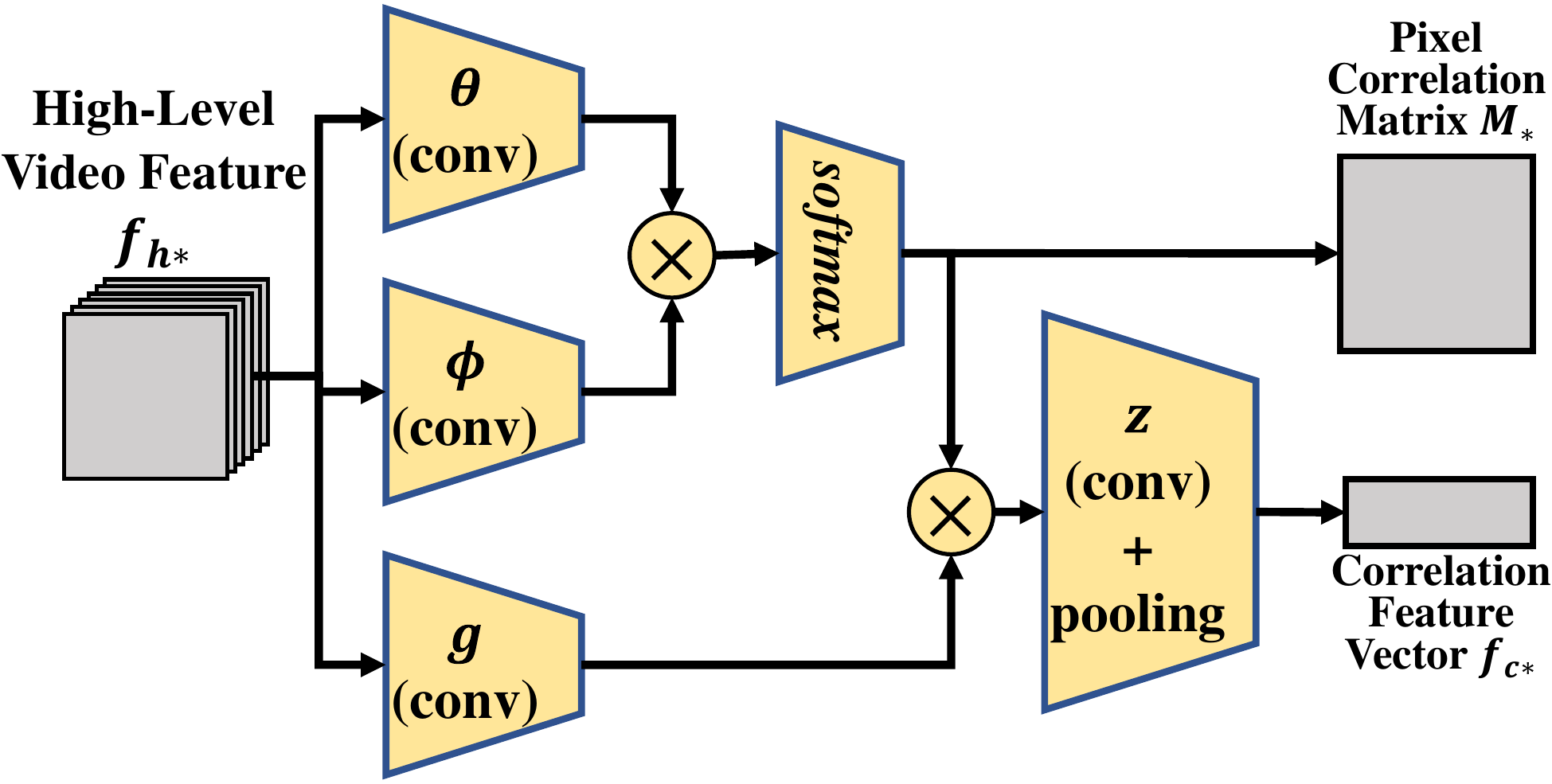}
\end{center}
   \caption{Structure of the correlation extraction module $G_c$. $G_c$ extract correlation features (pixel correlation matrix $\mathbf{M}_\ast$ and correlation feature vector $f_{c\ast}$) through the high-level video feature $f_{h\ast}$. It is built upon the non-local operation. $\mathbf{M}_\ast$ is obtained through multiplication of $f_{h\ast}$ projected on latent spaces, and represents the correlation between each spatiotemporal pixel feature. $f_{c\ast}$ is further obtained by multiplying the $\mathbf{M}_\ast$ $f_{h\ast}$ projected on the latent space, followed by pooling operation over spatiotemporal dimensions. The projection functions are implemented with convolution layers of $1\times1\times1$ kernel.}
\label{figure:3-2-corr}
\end{figure}

\subsection{Minimizing Pixel Correlation Discrepancy}
\label{section:method:pcd}
In the ACAN-Base network, the same DA approach is applied to both video and correlation features. However, it remains a question \textit{whether such an approach is the most effective way for aligning correlation features across different domains?} Aligning correlation features can be further achieved through aligning the joint distribution of correlation information. The joint distribution could be computed as the covariance of correlation information, implemented as its corresponding Gram matrix. The key to the above question therefore lies in the expression of the correlation information. As illustrated in Figure~\ref{figure:3-1-acan}, correlation features are extracted from $G_c$, whose structure is shown in Figure~\ref{figure:3-2-corr}. For the $i^{th}$ input video, we define the pixel correlation matrix (PCM) $\mathbf{M}_{\ast}^i$ as:
\begin{equation}
    \mathbf{M}_{\ast}^i = \varphi(\theta(f_{h\ast}^i)^T\phi(f_{h\ast}^i)),
\end{equation}
where $\varphi$ is the softmax operation. Both $\theta(\cdot)$ and $\phi(\cdot)$ are linear functions projecting the high-level video features to latent spaces. In practice, they are implemented as convolution layers with a kernel size of $1\times1\times1$. The value $\mathbf{M}_{\ast,pq}^i$ at the $(p,q)$ position of PCM represents the correlation between the video feature at spatiotemporal point p, $f_{h\ast,p}^i$, and the video feature at spatiotemporal point q, $f_{h\ast,q}^i$. We argue that PCM could be viewed as the correlation information of the video. Therefore the joint correlation information distribution is constructed as the Gram matrix of the PCM, denoted as $\mathcal{G}^i\in\mathbb{R}^{N_M\times N_M}$, where $N_M$ is the number of spatiotemporal points in the feature map $\theta(f_{h\ast}^i$). $\mathcal{G}^i$ is computed by:
\begin{equation}
    \mathcal{G}^i = {\mathbf{M}_{\ast}^i}^T \mathbf{M}_{\ast}^i.
\end{equation}

The alignment of correlation features thus requires the minimization of the distance between the Gram matrices $\mathcal{G}$, termed as the video covariance loss $\mathcal{L}_{vs}$, formulated by:
\begin{equation}
    \mathcal{L}_{vs} = \parallel\mathbf{E}(\mathcal{G}_{s}) - \mathbf{E}(\mathcal{G}_{t})\parallel^2,
\end{equation}
where the subscripts $s$ and $t$ denotes the Gram matrices for source and target videos respectively. However, such computation is inefficient, requiring a cost of $O({N_M}^2)$. Furthermore, improving network transferability through minimizing $\mathcal{L}_{vs}$ comes at the price of decreasing network discriminability. To minimize $\mathcal{L}_{vs}$ more efficiently while causing less impact on the network's discriminability, we simplify according to the theory in~\cite{li2017demystifying}.
\begin{theorem}
Given the Gram matrices $\mathcal{G}_{s}, \mathcal{G}_{y}$ constructed from source and target features $\mathbf{M}_{s}$, $\mathbf{M}_{t}$, the minimization of distance between the Gram matrices $\mathcal{L}_{vs}$ can be seen as a distribution alignment process from $\mathbf{M}_{t}$ to $\mathbf{M}_{s}$.
\end{theorem}

As proven in~\cite{li2017demystifying}, the above theorem indicates that minimizing $\mathcal{L}_{vs}$ could be reformulated as minimizing the distribution discrepancy of $\mathbf{M}_{t}$ and $\mathbf{M}_{s}$. Set the underlying distributions of $\mathbf{M}_{s}$ be $p_{Ms}$ and that of $\mathbf{M}_{t}$ be $p_{Mt}$. Here we propose the \textit{pixel correlation discrepancy} (PCD), denote as $d_M(p_{Ms},p_{Mt})$. Computing and minimizing this discrepancy is achieved by representing the distributions $p_{Ms}$ and $p_{Mt}$ as elements on the reproducing kernel Hilbert space (RKHS). As such, the distribution discrepancy could be defined as distance of distribution embedded elements on the RKHS.

Further, to align the distributions of $p_{Ms}$ and $p_{Mt}$ in a more fine-grained manner, it is important to align the distributions taking the relations between relevant classes into consideration. That is to align $p_{Ms}$ and $p_{Mt}$ within the same action classes in source and target domains, instead of aligning it only in by the global distributions. The overall PCD is therefore formulated as:
\begin{equation}
\label{eqn:method:pcd-1}
    d_{M}(p_{Ms},p_{Mt})\triangleq\mathbf{E}_{cl}\parallel\mathbf{E}_{p_{Ms}(cl)}[\zeta(\mathbf{M}_{s})] - \mathbf{E}_{p_{Mt}(cl)}[\zeta(\mathbf{M}_{t})]\parallel^2_{\mathcal{H}},
\end{equation}
where $\mathbf{E}_{p_{M\ast}(cl)}$ is the mean embedding of distribution $p_{M\ast}$ for action class $cl$ on the RKHS $\mathcal{H}$. The feature map $\zeta$ is closely related to the RKHS characteristic kernel $k$ by $k(\mathbf{M}_{s},\mathbf{M}_{t}) = \langle\zeta(\mathbf{M}_{s}),\zeta(\mathbf{M}_{t})\rangle$. The use of mean embedding for each class enables our PCD to align distributions of correlation information within each action class instead of only focusing on the global correlation information distribution. In practice, we may further assume that each video belongs to a certain action class with a class-related weight $w_{cl}$. We therefore could estimate PCD in Equation~\ref{eqn:method:pcd-1} as:
\begin{equation}
\label{eqn:method:pcd-2}
    \resizebox{0.9\linewidth}{!}{
    $d_{M}(p_{Ms},p_{Mt}) = \frac{1}{Cl}\sum\limits_{cl=1}^{Cl}\parallel\sum\limits_{i=1}^{N_s}w_{scl}^i\zeta(\mathbf{M}_{s}^i) - \sum\limits_{j=1}^{N_t}w_{tcl}^j\zeta(\mathbf{M}_{t}^j)\parallel^2_{\mathcal{H}}$,
    }
\end{equation}
where $Cl$ is the number of action classes. When computing the weight of a source video for a certain action class, given that the labels are provided, the weight $w_{scl}^i$ is computed by:
\begin{equation}
\label{eqn:method:source-weight}
    w_{scl}^i = \frac{y_s^i}{\sum\limits_{k=1}^{N_s}y_s^k}.
\end{equation}
Whereas for the target videos, since the labels are not available, we cannot compute the weight $w_{tcl}^j$ directly. Instead, we utilize the output from the action classifier $G_y$ which characterizes the probability of assigning a given video to an action class. This is denoted as the pseudo-label for a target video and is computed by:
\begin{equation}
\label{eqn:method:pseudo-label}
    y_t^j = G_{y}(f_t^j\oplus f_{ct}^j).
\end{equation}
The resulting pseudo-labels of the target videos could be used as in Eqn.~\ref{eqn:method:source-weight} for computing the weight of a target video for an action class. Finally, since the feature map $\zeta$ cannot be computed directly in most cases, we expand Eqn.~\ref{eqn:method:pcd-2} while utilizing the characteristic kernel k. The PCD could therefore be reformulated as:
\begin{equation}
\label{eqn:method:pcd-final}
    \begin{split}
    d_{M}(p_{Ms},p_{Mt}) = \frac{1}{C}&\sum\limits_{c=1}^{C}(  \sum\limits_{i=1}^{N_s}\sum\limits_{i'=1}^{N_s}w_{sc}^i w_{sc}^{i'} k(\mathbf{M}_{s}^i, \mathbf{M}_{s}^{i'}) \\
    & + \sum\limits_{j=1}^{N_t}\sum\limits_{j'=1}^{N_t}w_{tc}^i w_{tc}^{j'} k(\mathbf{M}_{t}^j, \mathbf{M}_{t}^{j'}) \\
    & - 2\sum\limits_{i=1}^{N_s}\sum\limits_{j=1}^{N_t}w_{sc}^i w_{tc}^{j} k(\mathbf{M}_{s}^i, \mathbf{M}_{t}^{j}) ),
    \end{split}
\end{equation}
where the kernel $k$ would typically be of Gaussian form, hence $k(\mathbf{M}_{s}^i, \mathbf{M}_{t}^{j})=-exp(\frac{\parallel\mathbf{M}_{s}^i-\mathbf{M}_{t}^{j}\parallel^2}{2\sigma^2})$. The overall optimization objective is thus formulated as:
\begin{equation}
\label{eqn:method:overall-loss-final}
    \mathcal{L} = \mathcal{L}_y - (\lambda_v\mathcal{L}_{vd} + \lambda_r\mathcal{L}_{cd}) + \lambda_d d_{M},
\end{equation}
where $\lambda_d$ is the trade-off weight for the PCD. Minimizing our proposed PCD is superior in effective alignment of cross-domain correlation features thanks to its relatively solid theoretical motivation. While aligning video features could also be achieved by minimizing feature discrepancies directly through methods such as MMD~\cite{long2015learning}, CORAL~\cite{sun2016deep}, these discrepancies cannot measure the correlation difference between the source and the target domains as in PCD which matters to video DA. Therefore, applying MMD or CORAL for video feature alignment produces inferior performances than our proposed approach as illustrated in Section~\ref{section:experiments}. For inference, we follow the steps as indicated in Algorithm~\ref{algorithm:3-1-acan} and obtain the action recognition predictions for the unlabeled target domain videos. Note that the video indices $i,j$ are omitted for simplicity.

\begin{algorithm}
	\caption{Inference ACAN for Target Domain Videos}
	\begin{algorithmic}
        \STATE \textbf{Input:} Target data $V_t\in \mathcal{D}_t$, trained feature generators $G_f$, $G_c$, and trained classifier $G_f$
        \STATE \textbf{Output:} Predicted action class $y_t$
		\STATE Obtain target video feature $f_t = G_f (V_t)$
		\STATE Obtain target correlation feature vector $f_{ct} = G_c (V_t)$
		\STATE Concatenate $f_t$ and $f_{ct}$ to form the overall feature representation of $V_t$ by $f_t\oplus f_{ct}$
		\STATE $y_t = G_{y}(f_t\oplus f_{ct})$
	\end{algorithmic}
\label{algorithm:3-1-acan}
\end{algorithm}

%------------------------------------------------------------------------
\section{The HMDB-ARID Dataset}
\label{section:dataset}

\begin{table}[t]
\centering
\caption{Comparison of RGB mean and standard deviation (std) over common action recognition datasets and the ARID dataset.}
\resizebox{.8\linewidth}{!}{
\begin{tabular}{l|c|c}
\hline
\hline
Dataset & RGB Mean & RGB Std \Tstrut\\
\hline
HMDB51 & [0.424,0.364,0.319] & [0.268,0.255,0.260]\Tstrut\\
UCF101 & [0.409,0.397,0.358] & [0.266,0.265,0.270]\\
Kinetics & [0.432,0.395,0.377] & [0.228,0.222,0.217]\\
\hline
ARID & [0.079,0.074,0.073] & [0.101,0.098,0.090]\Tstrut\\
\hline
\hline
\end{tabular}
}
\label{table:4-1-dataset}
\end{table}

\begin{table*}[t]
\centering
\caption{Comparison of current and our novel video DA datasets.}

\makebox[0pt]{
\resizebox{.85\linewidth}{!}{
\begin{tabular}{c|c|c|c|c}
\hline
\hline
Statistics & UCF-HMDB\textsubscript{\textit{small}} & UCF-Olympic & UCF-HMDB\textsubscript{\textit{full}} & HMDB-ARID \Tstrut\\
\hline
Video Length (seconds) & 1-21 & 1-39 & 1-33 & 1-30 \Tstrut\\
Video Classes \# & 5 & 6 & 12 & 11\\
Training Video \# & UCF:482/HMDB:350 & UCF:601/Olympic:250 & UCF:1438/HMDB:840 & HMDB:770/ARID:2288 \\
Validation Video \# & UCF:189/HMDB:150 & UCF:240/Olympic:54 & UCF:571/HMDB:360 & HMDB:330/ARID:823 \\
\hline
\hline
\end{tabular}
}
}
\label{table:4-2-cross_comparison}
\end{table*}

There are very limited cross-domain benchmark datasets for video DA tasks, therefore hindering the research for video DA. Previous cross-domain datasets introduced for video DA~\cite{sultani2014human,xu2016dual,jamal2018deep} are of very small-scale, with not more than 6 classes, and typically less than 1,000 videos. The lack of classes and data over these cross-domain datasets introduces limited domain discrepancy, and therefore the performances of DA approaches are saturated. More recently, larger cross-domain video datasets, such as UCF-HMDB\textsubscript{\textit{full}} have been introduced with larger domain discrepancies.

Though larger cross-domain datasets have been introduced, both domains included in these datasets are still based on current well-established action recognition datasets. These action recognition datasets may include different classes with different videos, yet most of them are collected on public video platforms. This would lead to similar video statistics among these datasets, as compared in Table~\ref{table:4-1-dataset}. Similar video statistics suggest high probability of similar scenarios exist among current action recognition datasets, thus the domain shift between these datasets may not be significant. Consequently, the difficulty of adapting the same model across the different domains with similar video statistics or similar scenarios may be trivial. Video DA approaches that perform well in these cross-domain video datasets may not be well applicable in real-world applications where the gap between domains may be much larger than current cross-domain datasets. We argue that video DA approaches would be more useful for bridging with video domains with large distribution shifts, such as dark videos (adverse illumination) or hazy videos (adverse contrast).

\begin{figure*}[t]
\begin{center}
   \includegraphics[width=.75\linewidth]{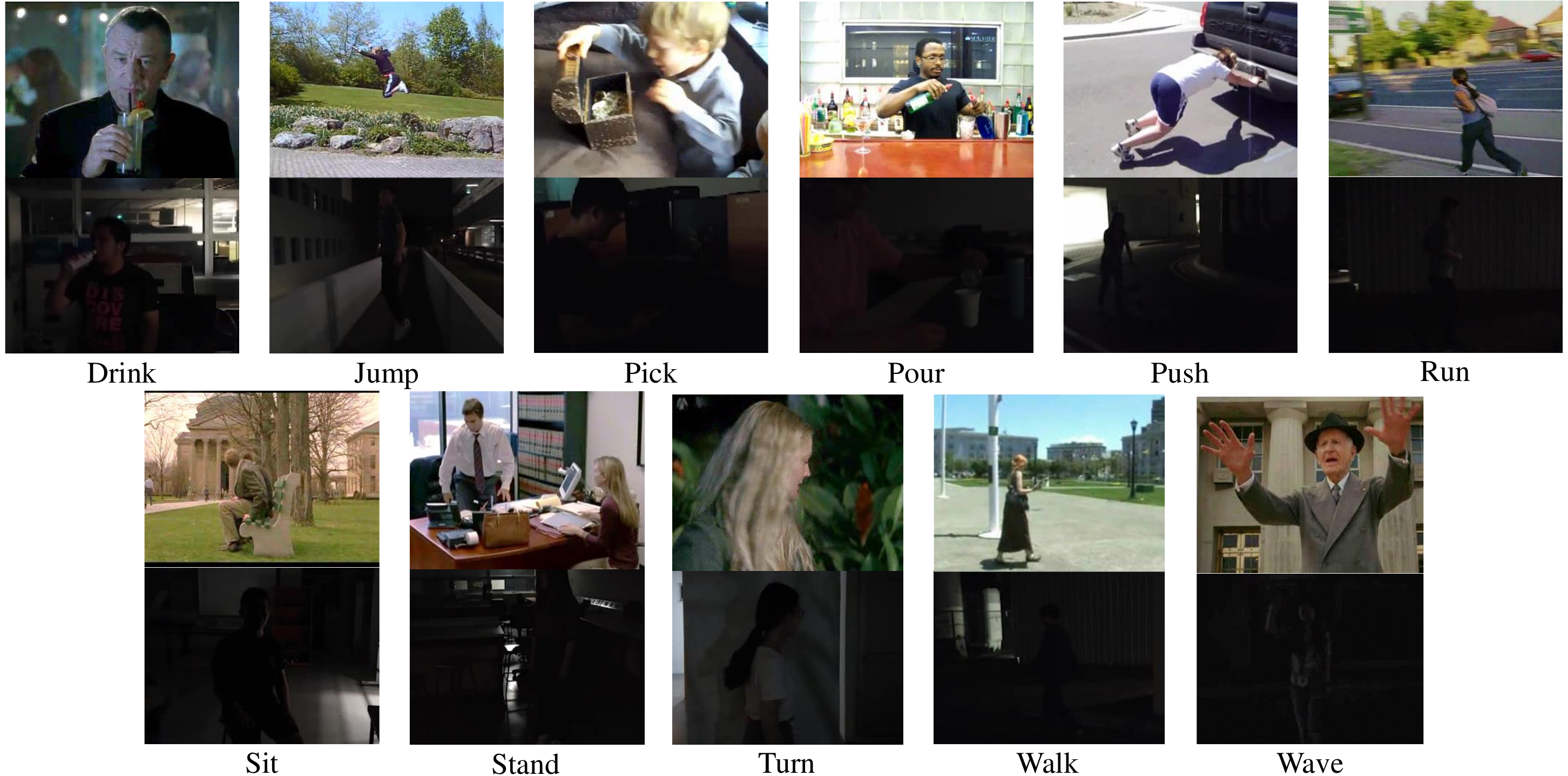}
\end{center}
   \caption{Sampled frames for each action class from the videos in HMDB-ARID. Note that the sampled frames from HMDB51 are shown in the upper row, whereas the sampled frame from ARID are shown in the lower row. Best viewed zoomed in.}
\label{figure:supp-2-sample}
\end{figure*}

To explore how to leverage current datasets to boost performance on videos shot in adverse environments, we propose a novel cross-domain dataset. It incorporates both the current action recognition dataset and a more recent dark dataset, ARID~\cite{xu2020arid}, whose videos are shot under adverse illumination conditions. Compared with current action recognition datasets, videos in ARID are characterized by low brightness and low contrast. Statistically, videos in ARID possess much lower RGB mean value and standard deviation (std), as presented in Table~\ref{table:4-1-dataset}. The larger statistical differences between ARID and current action recognition datasets, such as HMDB51~\cite{kuehne2011hmdb}, would strongly suggest a larger domain shift between the different datasets.

The ARID dataset includes a total of 11 human action classes. These includes \textit{drink, jump, pick, pour, push, run, sit, stand, turn, walk and wave}. When proposing the cross-domain HMDB-ARID dataset, we include all 11 action classes in ARID and HMDB51. For both datasets, we follow the official split method to separate the train and validation sets. The HMDB-ARID dataset thus includes 770 training videos and 330 validation videos from HMDB51, and 2288 training videos and 823 validation videos from ARID. Figure~\ref{figure:supp-2-sample} shows the comparison of sampled frames from HMDB-ARID dataset. Compared to previous video DA datasets, besides containing larger domain shift, our dataset also contains a larger number of total videos for both training and validation, as illustrated in Table~\ref{table:4-2-cross_comparison}.

%------------------------------------------------------------------------
\section{Experiments}
\label{section:experiments}

In this section, we evaluate our proposed ACAN performing cross-domain action recognition on two video DA datasets: UCF-HMDB\textsubscript{\textit{full}} and our new HMDB-ARID. We present state-of-the-art results on both datasets. We also present detailed ablation studies and qualitative analysis of our proposed ACAN to verify our design.

\subsection{Experimental Settings and Details}
\label{section:experiment:settings}
We perform action recognition tasks on both the UCF-HMDB\textsubscript{\textit{full}} dataset and our new HMDB-ARID dataset. The UCF-HMDB\textsubscript{\textit{full}} dataset~\cite{chen2019temporal} is introduced as an expansion of the original UCF-HMDB\textsubscript{\textit{small}} dataset~\cite{sultani2014human}, with more classes and larger domain discrepancy. The UCF-HMDB\textsubscript{\textit{full}} contains a total of 3,209 videos with 12 action classes, all from the original UCF101~\cite{soomro2012ucf101} and HMDB51~\cite{kuehne2011hmdb} datasets. It includes two settings: UCF$\to$HMDB and HMDB$\to$UCF, where the direction of the arrow symbol is set from the source domain towards the target domain. We use the same splits as provided in the original paper~\cite{chen2019temporal}. The novel HMDB-ARID dataset is as introduced in Section~\ref{section:dataset}, and also consist of two settings: HMDB$\to$ARID and ARID$\to$HMDB. For all four settings, we report the top-1 accuracy on the target dataset, averaged on 5 runs with identical settings for each approach.

Our experiments are implemented using the PyTorch~\cite{paszke2019pytorch} library. To obtain video features, we instantiate two 3D-CNNs, I3D~\cite{carreira2017quo} and MFNet~\cite{chen2018multi}, as $G_f$ for both source and target domain videos. Both I3D and MFNet are utilized thanks to its performance on current action recognition benchmarks (namely UCF101~\cite{soomro2012ucf101}, HMDB51~\cite{kuehne2011hmdb} and Kinetics~\cite{kay2017kinetics}). MFNet is also utilized due to its lightweight structure, which enables it to achieve comparable results to that of I3D while requiring a fraction of the parameters and computation power needed.

The source and target feature extractors share parameters. 
% \textcolor{blue}{
Following the implementation in~\cite{chen2018multi,carreira2017quo}, the input for both I3D and MFNet as the source or target feature extractors are frame sequences of 16 frames sampled sequentially from the original input source or target video. Each frame is of the same resolution obtained by resizing such that the shorter edge is of 240 pixels and cropping the original frame to resolution $224\times224$.
% }
The correlation extraction module takes the high-level video feature from the output of $layer4$ in I3D and the output of $conv4$ layer in MFNet as inputs, which are feature maps of size $14\times14$. The stochastic gradient descent algorithm~\cite{bottou2010large} is used for optimization, with the weight decay set to 0.0001 and the momentum to 0.9 for both I3D and MFNet. During training, the batch size is set to 8 samples per GPU. Empirically, Our initial learning rate is set to 0.005 and is divided by 10 after 20 and 35 epochs. $\lambda_v$ is set to 0.5 while $\lambda_r$ and $\lambda_d$ are both set to 1.0 through empirical results. All experiments are conducted using two NVIDIA GP100 GPUs.

\subsection{Overall Results}
\label{section:experiment:results-comparion}
There are limited studies focusing on applying DA approaches to the action recognition task. Here we first compare previous methods utilizing the UCF-HMDB\textsubscript{\textit{full}} benchmark. These include TA\textsuperscript{3}N~\cite{chen2019temporal}, TCoN~\cite{pan2020adversarial} and SAVA~\cite{choi2020shuffle}. Due to the different encoders used for the different methods, we report both (a) the ``Source only" results, where the network is trained with supervised source data only and validated on the target data, and is the lower bound performance for the adaptation process; and (b) the ``Target only" results, where the network is directly trained and validated with supervised target data and is the upper bound performance for the adaptation process. The comparison of performance should focus on the networks' improvement with respect to the performance with the ``Source only" setting. The comparison should also focus on the distance between the network's performance and the performance with the ``Target only" setting. For the performance of TA\textsuperscript{3}N. we follow the works in~\cite{choi2020shuffle} and obtain the results by running the publicly available code. Table~\ref{table:5-1-compare-u_h} shows the comparison of performances between our proposed ACAN and the methods as mentioned on UCF-HMDB\textsubscript{\textit{full}}.

\begin{table}[t]
\centering
\caption{Results on the two settings for UCF-HMDB\textsubscript{\textit{full}}}
\resizebox{.95\linewidth}{!}{
\begin{tabular}{l|c|c|c}
\hline
\hline
Method & Encoder & UCF$\to$ HMDB & HMDB$\to$ UCF \Tstrut\\
\hline
Source Only & TRN-Res101 & 73.1\% & 73.9\% \Tstrut\\
TA\textsuperscript{3}N & TRN-Res101 & 75.3\% & 79.3\%\\
TCoN & TRN-Res101 & \textbf{87.2\%} & 89.1\%\\
Target Only & TRN-Res101 & 90.8\% & 95.6\%\\
\hline
Source Only & I3D & 80.3\% & 88.8\% \Tstrut\\
SAVA & I3D & 82.2\% & 91.2\%\\
\textbf{ACAN}(Ours) & I3D & 85.4\% & \textbf{93.8\%}\\
Target Only & I3D & 95.0\% & 96.8\%\\
\hline
Source Only & MFNet & 78.6\% & 88.4\% \Tstrut\\
\textbf{ACAN}(Ours) & MFNet & \textbf{85.8\%} & 93.2\%\\
Target Only & MFNet & 96.0\% & 97.1\%\\
\hline
\hline
\end{tabular}
}
\label{table:5-1-compare-u_h}
\end{table}

\begin{table}[t]
\centering
\caption{Results on the two settings for HMDB-ARID.}
\resizebox{.95\linewidth}{!}{
\begin{tabular}{l|c|c|c}
\hline
\hline
Method & Encoder & HMDB$\to$ ARID & ARID$\to$ HMDB \Tstrut\\
\hline
Source Only & TRN-Res101 & 17.8\% & 15.7\% \Tstrut\\
TA\textsuperscript{3}N & TRN-Res101 & 22.4\% & 19.8\%\\
Target Only & TRN-Res101 & 52.8\% & 50.9\%\\
\hline
Source Only & MFNet & 48.3\% & 37.9\% \Tstrut\\
DANN & MFNet & 50.7\% & 40.6\%\\
MK-MMD & MFNet & 50.2\% & 40.1\%\\
MCD & MFNet & 47.6\% & 36.8\%\\
CORAL & MFNet & 51.3\% & 41.7\%\\
\textbf{ACAN}(Ours) & MFNet & \textbf{58.0\%} & \textbf{46.4\%}\\
Target Only & MFNet & 76.1\% & 67.6\%\\
\hline
\hline
\end{tabular}
}
\label{table:5-2-compare-h_a}
\end{table}

The performance results in Table~\ref{table:5-1-compare-u_h} shows that our proposed ACAN achieves the best result under the HMDB$\to$UCF setting and very competitive performance under the UCF$\to$HMDB setting when using either MFNet or I3D as the encoder. More specifically, our ACAN with the MFNet encoder achieves $85.8\%$ top-1 accuracy for UCF$\to$HMDB setting, indicating that the improvement brought by ACAN towards the lower bound of the UCF$\to$HMDB setting is $7.2\%$. This is significantly higher than that brought by SAVA ($1.9\%$) and TA\textsuperscript{3}N ($2.2\%$). The large improvement brought by ACAN enables our network to perform better on UCF$\to$HMDB setting despite the lower bound of MFNet is lower than that of I3D~\cite{carreira2017quo}. Under this setting, our ACAN is also closer to the upper bound of the encoder, with a gap of $10.2\%$. Comparatively, the gap to the upper bound performance is $15.5\%$ for TA\textsuperscript{3}N and $12.8\%$ for SAVA. Similarly, our ACAN with I3D encoder also performs better than both TA\textsuperscript{3}N and SAVA. Comparatively, ACAN with I3D encoder outperforms SAVA by $3.2\%$ while sharing the I3D as the common video feature encoder with SAVA. This further demonstrates the superiority of ACAN over current video DA methods.

The superiority of ACAN further strengthens under the HMDB$\to$UCF setting. Under this settings when utilizing MFNet as the video feature encoder, our proposed ACAN gains a $4.8\%$ improvement towards the lower bound performance, which is greater than that brought by SAVA ($2.4\%$). When utilizing I3D as the video feature encoder as in SAVA, our proposed ACAN gains an exceptional $5.0\%$ improvement towards the lower bound performance. The larger increase built upon the strong I3D encoder enables our ACAN to achieve the best result under this setting with $93.8\%$ top-1 accuracy. The gap towards the upper bound performance is also the smallest for ACAN using the I3D encoder, with $3.0\%$ compared to $16.3\%$ for TA\textsuperscript{3}N, $6.5\%$ for TCoN, and $5.6\%$ for SAVA.

We further compare performances of several methods on our novel HMDB-ARID dataset, with both HMDB$\to$ARID and ARID$\to$HMDB settings, as shown in Table~\ref{table:5-2-compare-h_a}. Note that both settings are more challenging, given that the gap between the lower bound performance (trained with supervised source data) and the upper bound performance (trained with supervised target data) is larger compared to the settings for UCF-HMDB\textsubscript{\textit{full}}. In addition to comparing with the TA\textsuperscript{3}N with TRN-Res101~\cite{zhou2018temporal} encoder, we also compare with performances with other typical DA approaches, e.g. DANN~\cite{ganin2015unsupervised}, MK-MMD~\cite{long2015learning}, MCD~\cite{saito2018maximum}, and CORAL~\cite{sun2016deep}, all with MFNet as the encoder.

The performance results in Table~\ref{table:5-2-compare-h_a} indicate that our proposed ACAN achieves the best results in either setting related to our novel HMDB-ARID dataset. Our ACAN achieves a top-1 accuracy of $58.0\%$ for the HMDB$\to$ARID setting and $46.4\%$ for the ARID$\to$HMDB setting. Our ACAN also brings the most significant improvement with respect to the lower bound performance, with $9.8\%$ and $8.5\%$ for the two settings respectively. Comparatively, TA\textsuperscript{3}N which does not utilize correlation alignment only brings $4.6\%$ and $4.1\%$ increase with respect to the lower bound performance. This shows that previous methods that fail to align correlations would not be able to effectively handle the larger domain shift caused by a more significant difference in video statistics. Note that the gap to the upper bound performance obtained by training with supervised target data is still relatively large, suggesting further improvements could be made on this novel HMDB-ARID dataset.

\subsection{Ablation Studies}
\label{section:experiment:ablation}
We further justify our proposed design of ACAN through thorough ablation studies. Specifically, we first examine the performance of our ACAN in four scenarios and justify the need for introducing correlation features in the extraction process, the use of two separate domain losses, and the introduction of PCD. We also introduce an alternative form of the joint correlation information distribution difference minimization to compare and justify our current design of PCD. All ablation studies are conducted under the UCF$\to$HMDB and HMDB$\to$ARID settings, with the batch size and other training parameters as mentioned in Section~\ref{section:experiment:settings}. The MFNet~\cite{chen2018multi} is instantiated as the encoder for all ablation studies.

\begin{figure*}[t]
\begin{center}
   \includegraphics[width=.9\linewidth]{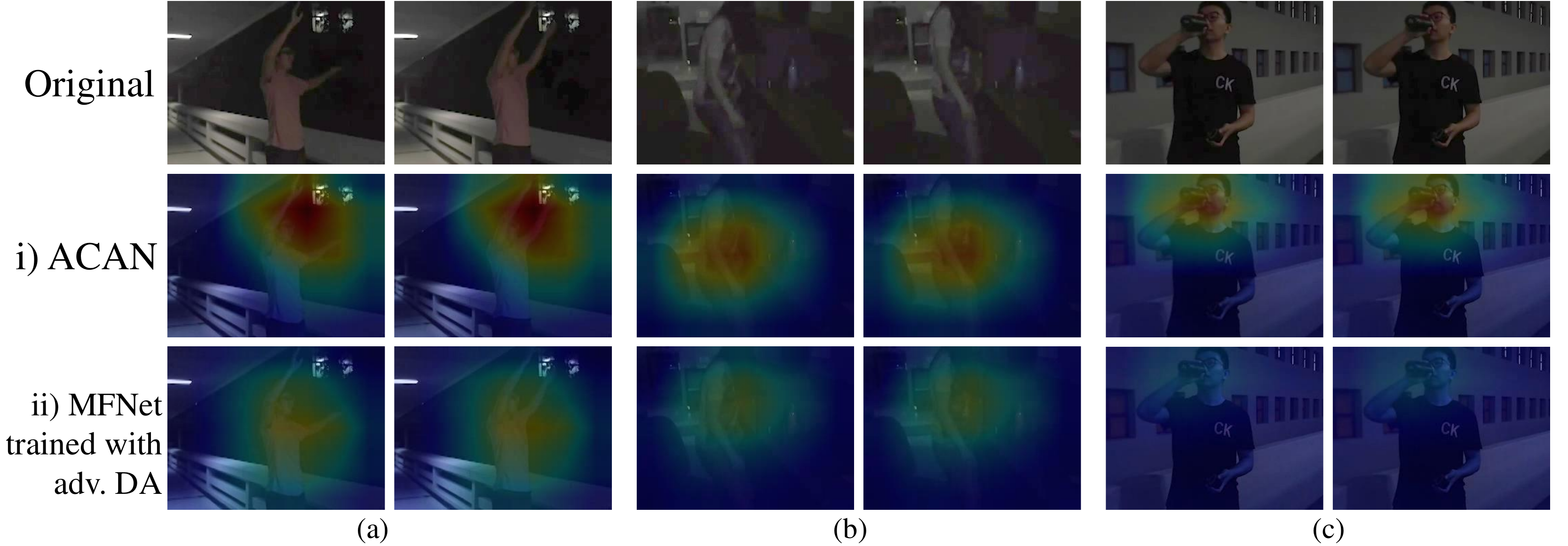}
\end{center}
   \caption{Class activation maps (CAMs) on ARID, utilizing i) ACAN and ii) MFNet trained with adversarial DA approach. The CAMs are obtained from three actions: (a) ``Wave"; (b) ``Stand"; and (c) ``Drink". We also show the original frames at the top row from which the CAMs are computed. The original frames are tuned brighter for visualization.}
\label{figure:5-1-cam}
\end{figure*}

\begin{figure}[t]
\begin{center}
   \includegraphics[width=1.\linewidth]{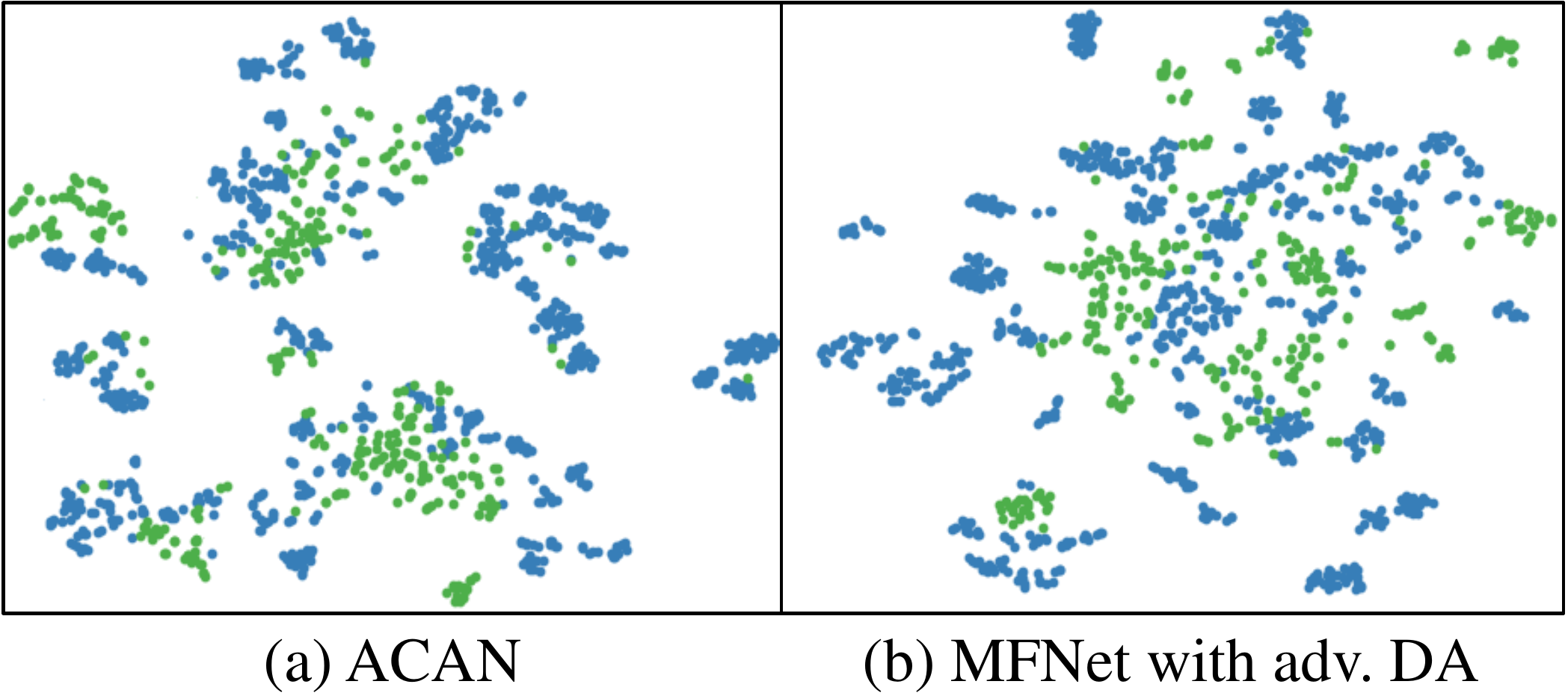}
\end{center}
   \caption{Comparison of t-SNE visualization of video features of both source and target domains under HMDB$\to$ARID. The video features are obtained from (a) ACAN and (b) MFNet trained with the adversarial DA approach. The green dots represent the data from the source domain while the blue dots represent the data from the target domain.}
\label{figure:5-2-tsne}
\end{figure}

\begin{table}[t]
\centering
\caption{Ablation experiments on including correlation features, on UCF$\to$HMDB and HMDB$\to$ARID settings.}
\resizebox{1.\linewidth}{!}{
\begin{tabular}{l|c|c}
\hline
\hline
Method & UCF$\to$HMDB & HMDB$\to$ARID \Tstrut\\
\hline
Source only w/o. correlation & 76.1\% & 48.1\% \Tstrut\\
Source only w. correlation & 78.6\% & 48.3\%\\
\hline
Adv. DA w/o. correlation & 80.2\% & 50.7\% \Tstrut\\
Adv. DA w. correlation & 84.2\% & 52.6\%\\
\hline
\hline
\end{tabular}
}
\label{table:5-3-ablation-1}
\end{table}

\begin{table}[t]
% \small
\centering
\caption{Ablation experiments on the domain loss $\mathcal{L}_d$ on UCF$\to$HMDB and HMDB$\to$ARID settings.}
\resizebox{.75\linewidth}{!}{
\begin{tabular}{l|c|c}
\hline
\hline
Method & UCF$\to$HMDB & HMDB$\to$ARID \Tstrut\\
\hline
ACAN & 85.8\% & 58.0\% \Tstrut\\
\hline
ACAN$-\mathcal{L}_{cd}$ & 84.9\% & 56.9\% \Tstrut\\
ACAN$-\mathcal{L}_{vd}$ & 84.5\% & 56.7\%\\
\hline
MFNet + PCD & 83.8\% & 56.1\% \Tstrut\\
\hline
\hline
\end{tabular}
}
\label{table:5-4-ablation-2}
\end{table}

\begin{table}[t]
\centering
\caption{Ablation on PCD and alternative way of minimizing joint correlation information distribution difference, on UCF$\to$HMDB and HMDB$\to$ARID settings.}
\resizebox{.8\linewidth}{!}{
\begin{tabular}{l|c|c}
\hline
\hline
Method & UCF$\to$HMDB & HMDB$\to$ARID \Tstrut\\
\hline
ACAN & 85.8\% & 58.0\% \Tstrut\\
\hline
ACAN-Base & 84.2\% & 52.6\% \Tstrut\\
ACAN (l2-norm) & 85.0\% & 54.2\%\\
\hline
\hline
\end{tabular}
}
\label{table:5-5-ablation-3}
\end{table}

\textbf{The necessity of correlation feature alignment.} We first justify the need for correlation features for alignment, which is achieved by (a) comparing the ``Source only" results with and without the introduction of correlation features, and (b) comparing the use of adversarial DA approaches with and without correlation features. Results in Table~\ref{table:5-3-ablation-1} justifies the use of correlation features, where such strategy consistently improves the performance of the network under both ``Source only" training and when DANN method is used for DA. It could also be observed that the use of correlation features brings more improvement when the DANN method is applied. Such observation is consistent with our argument of improving video feature alignment by using correlation alignment.

\textbf{The effectiveness of domain loss $\mathcal{L}_d$.} We then justify our design of the domain loss $\mathcal{L}_d$, which is the weighted sum of $\mathcal{L}_{vd}$ and $\mathcal{L}_{cd}$. We compare with the variants of ACAN where either $\mathcal{L}_{vd}$ or $\mathcal{L}_{cd}$ alone is used as the domain loss, denoted as ACAN$-\mathcal{L}_{cd}$ and ACAN$-\mathcal{L}_{vd}$. We also tested on the case where the domain loss is not applied (hence aligning correlation features by minimizing PCD alone), denoted as MFNet$+$PCD. As indicated in Table~\ref{table:5-4-ablation-2}, both losses contribute to the effective alignment of video features. The removal of either loss brings a decrease in network performance for both dataset settings. Further decrease is observed when no domain loss is applied. Meanwhile, the domain discriminators corresponding to either domain loss bring only a negligible growth in computation cost. Hence it is worthwhile to include two separate domain discriminators, with two domain losses for the overall domain loss $\mathcal{L}_d$.

\textbf{The effectiveness of PCD.} PCD is introduced for improving the effectiveness of correlation alignment by matching the joint correlation information distribution of video domains. We examine the effect of PCD through comparing with the ACAN variant without PCD, which is ACAN-Base as shown in Figure~\ref{figure:3-1-acan}. The results in Table~\ref{table:5-5-ablation-3} demonstrates the effectiveness of PCD, whose absence results in a noticeable $1.6\%$ accuracy decrease for UCF$\to$HMDB setting, and a significant $5.4\%$ accuracy decrease for HMDB$\to$ARID setting. Though the introduced PCD improves the effectiveness of correlation alignment greatly, minimizing PCD involves kernel estimation which increase computation cost. Inspired by the hypothesis presented in~\cite{xu2019larger}, minimizing the joint distribution difference, and hence the distance between distributions $p_{Ms}$ and $p_{Mt}$, could also be achieved through matching the norm of $p_{Ms}$ and $p_{Mt}$ towards a shared restrictive scalar $R$. The computation of distribution distance with this method is simpler given that no kernel estimation is required. In this case, the equation for the overall loss Equation~\ref{eqn:method:overall-loss-final} is reformulated as:
\vspace{0.5em}
\begin{equation}
\label{eqn:exp:loss-variant}
    \begin{split}
    \mathcal{L} = & \mathcal{L}_y - (\lambda_v\mathcal{L}_{vd} + \lambda_r\mathcal{L}_{cd}) +
    \\ &
    \lambda_{dist}(L_{dist}(\frac{1}{N_s}\sum\limits_{i=1}^{N_s}n(\mathbf{M}_s^i), R) +
    % \\ &
    L_{dist}(\frac{1}{N_t}\sum\limits_{j=1}^{N_t}n(\mathbf{M}_t^j), R)).
    \end{split}
\end{equation}
Here $L_{dist}$ is the distance loss between the norm of PCMs and the restrictive scalar $R$, and is implemented as $L_2$-distance, while $n(\cdot)$ denotes the norm function. $R$ is set to 25 during the experiments. We denote the variant of ACAN with loss function in Eqn.~\ref{eqn:exp:loss-variant} as ACAN (l2-norm) and compare with the original ACAN. The results in Table~\ref{table:5-5-ablation-3} shows that the variant formulated by Equation~\ref{eqn:exp:loss-variant} could still bring noticeable improvement compared to the ACAN-Base where the distributions of $p_{Ms}$ and $p_{Mt}$ are not aligned. However, compared to PCD, the improvement is relatively minor, which further justifies the effectiveness of the current design of PCD.

\subsection{Qualitative Analysis}
\label{section:experiment:qualitative}
To better understand the effect of ACAN, we perform qualitative analysis on trained networks. We first present the class activation maps (CAM)~\cite{zhou2016learning} of the target ARID videos with ACAN and with MFNet (encoder) trained with adversarial DA approach in Figure~\ref{figure:5-1-cam}. The dark videos in ARID make it difficult for accurate video features to be extracted. Therefore if correlation alignment is not utilized, the network may fail to focus on the actual action in the target domain. Instead, it may only briefly focus on the whole actor (Figure~\ref{figure:5-1-cam}(ii-a)), or on unrelated background (Figure~\ref{figure:5-1-cam}(ii-b)). With the involvement of correlation features and its alignment, ACAN is able to focus on the waving hand for the ``Wave" action, or the person standing for the ``Stand" action, thus showing much stronger performance on the HMDB$\to$ARID setting. Further, we visualize the distribution of the source and target domains under the HMDB$\to$ARID setting with t-SNE \cite{van2008visualizing}, as shown in Figure~\ref{figure:5-2-tsne}. It could be observed that our proposed ACAN can group both the data from the source domain (green dots) and data from the target domain (blue dots) into denser clusters. Our ACAN could also match the target domain data with source domain data more accurately.

%------------------------------------------------------------------------
\section{Conclusion and Future Work}
\label{section:conclusion}

In this work, we propose a novel domain adaptation method for action recognition across different domains. The new ACAN aligns correlation features in an adversarial manner while minimizing joint correlation information distribution differences by minimizing PCD. We further introduce a novel video DA dataset, HMDB-ARID, with a larger domain shift, and is the first video DA dataset that includes videos shot in adverse conditions. Our method obtains state-of-the-art results on both the UCF-HMDB\textsubscript{\textit{full}} and HMDB-ARID datasets. We further justify our design via thorough ablation studies and validate the effectiveness of ACAN with qualitative results.

Although state-of-the-art performances have been achieved by the proposed ACAN, we observe that the gap to the upper bound performance obtained by training with supervised target data is still relatively large as depicted in Table~\ref{table:5-2-compare-h_a}, suggesting further improvements could be made on the novel HMDB-ARID dataset. Additionally, cross-domain video datasets that involves a variety of large domain shift scenarios, such as blurry or hazy videos may be explored. Video DA approaches that cope with these different large domain shift scenarios would also be further investigated.

%------------------------------------------------------------------------
\section*{Acknowledgement}

This research is supported by A*STAR Singapore under its Career Development Award (Grant No.\ C210112046). This work is jointly supported by NTU Presidential Postdoctoral Fellowship, ``Adaptive Multimodal Learning for Robust Sensing and Recognition in Smart Cities'' project fund, in Nanyang Technological University, Singapore.

%------------------------------------------------------------------------
% \clearpage
% \newpage
\bibliographystyle{IEEEtran}
\bibliography{IEEEabrv,acan_tnnls}
%------------------------------------------------------------------------
%------------------------------------------------------------------------
\end{document}